\documentclass[runningheads]{llncs}
\usepackage{graphicx}
\begin{document}
\title{FollowMeUp Sports: New Benchmark for 2D Human Keypoint Recognition}
%
%
\author{Ying Huang$^*{}^\dagger{}$\inst{1} \and
Bin Sun$^*$\inst{2} \and
Haipeng Kan$^*$\inst{2} \and
Jiankai Zhuang\thanks{Equal contribution. $^\dagger$The work was done at Keep Inc. The research was partially supported by the National Key Research and Development Program of China (2017YFB1002803).}\inst{3} \and
Zengchang Qin\inst{3}}
\authorrunning{Y. Huang et al.}
%
\institute{Alibaba Business School, Hangzhou Normal University, Hangzhou, China \email{yw155@buaa.edu.cn} 
\and Keep Inc., Beijing, China \email{\{sunbin,kanhaipeng\}@keep.com} 
\and Intelligent Computing and Machine Learning Lab, School of ASEE, Beihang University, Beijing, China \email{\{zhuangjk, zcqin\}@buaa.edu.cn}
}
\maketitle              
\begin{abstract}
Human pose estimation has made significant advancement in recent years. However, the existing datasets are limited in their coverage of pose variety. In this paper, we introduce a novel benchmark "FollowMeUp Sports" that makes an important advance in terms of specific postures, self-occlusion and class balance, a contribution that we feel is required for future development in human body models. This comprehensive dataset was collected using an established taxonomy of over 200 standard workout activities with three different shot angles. The collected videos cover a wider variety of specific workout activities than previous datasets including push-up, squat and body moving near the ground with severe self-occlusion or occluded by some sport equipment and outfits. Given these rich images, we perform a detailed analysis of the leading human pose estimation approaches gaining insights for the success and failures of these methods.

\end{abstract}
\section{Introduction}
\label{sec:intro}
Human pose estimation is an important computer vision problem \cite{johnson2010clustered}. Its basic task is to find the posture of a person via recognising human joints and rigid parts from normal RGB images. The extracted pose information is essential to modelling and understanding the human behaviours, and can be used in many vision application problems, such as virtual/augmented reality, human-computer interaction, action recognition and smart perception.

\begin{figure}[t]
\centering
\includegraphics[scale=1]{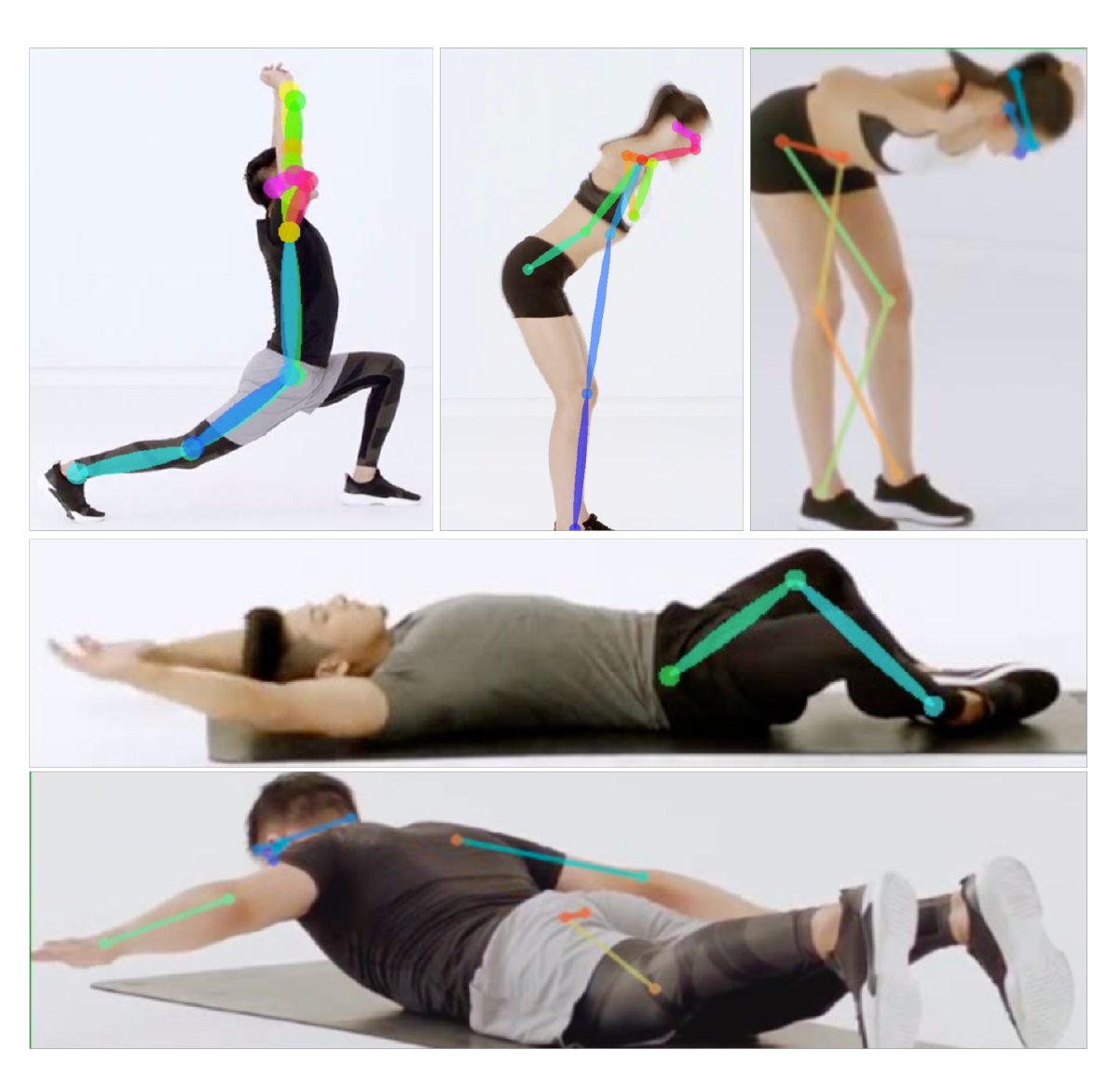}
\caption{Limitations of applying current pose estimation models on some workout postures, which have severe self-occlusion. Some body keypoints are falsely detected or missed in prediction even the background is plain.}
\label{falseSamples}
\end{figure}

In the psst few years, pose estimation methods based on deep neural network techniques have achieved great progress\cite{wei2016convolutional}\cite{newell2016stacked}\cite{luvizon20182d}. Although the performance of some human pose estimation models (e.g. \cite{chu2016structured}\cite{chu2017multi}\cite{sun2019deep}) is almost saturated on the above mentioned datasets, applying these high-precision algorithms to the other specific industrial tasks shows a degradation in accuracy. For instance, one application case is workouts or sports scoring. In this case, lots of activities have severe self-occlusion or unusual postures, such as push-up and crunch. We find out the models \cite{he2017mask}\cite{cao2017realtime}\cite{osokin2018real} trained on the MS-COCO dataset \cite{lin2014microsoft} cannot correctly detect body joints with atypical postures, as shown in Fig.~\ref{falseSamples}. In the top-right image of Fig.~\ref{falseSamples}, the right knee is falsely detected as left knee. In the top-left and lower-part images of Fig.~\ref{falseSamples}, some body joints, such as shoulders, knees and ankles, are missed in prediction. Since the pose estimation results of the same person in the standing posture are correct, we argue the false predictions are caused by the abnormal postures. Current datasets lack the corresponding samples\cite{andriluka20142d}\cite{andriluka2018posetrack}.

\begin{figure}[t]
\centering
\includegraphics[scale=0.2]{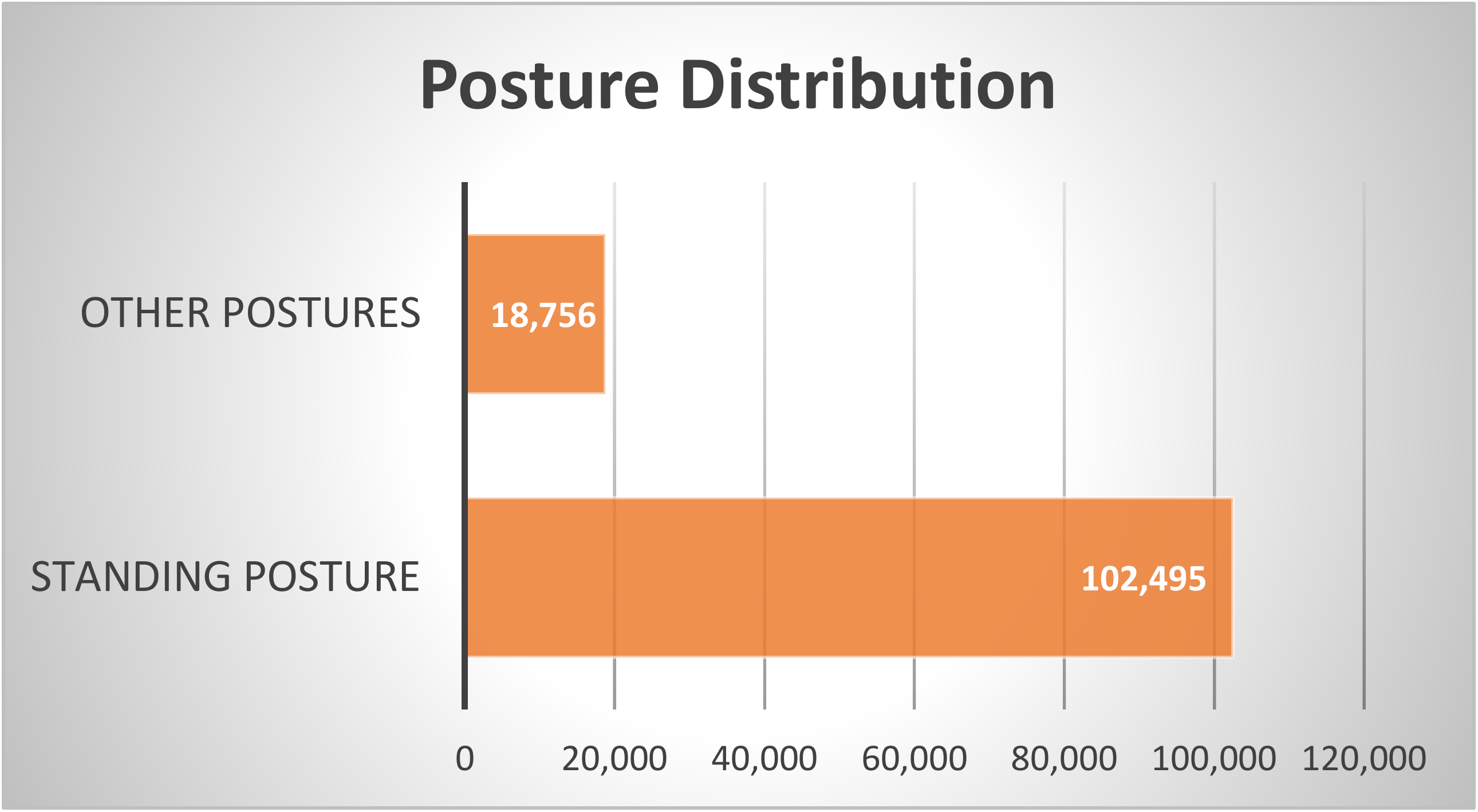}
\caption{The posture distribution of MS-COCO dataset. Around 85\% human instances are standing with good, upright posture.}
\label{figSampleDistribution}
\end{figure}

We use the MS-COCO dataset \cite{lin2014microsoft} as an example to analyse the distribution of human postures. In our statistics, the number of human instances in standing posture achieves 102,495 (84.53\%) while people in other postures only have 18,756 (15.47\%) as shown in Fig. \ref{figSampleDistribution}. The human instances in a horizontal position or an uncommon pose are extremely rare. This makes the model unable to learn the knowledge of irregular postures during training.

To improve the performance of human pose estimation in the certain sports situation, a large-scale human keypoints benchmark is presented in this paper. Our benchmark significantly advances state-of-the-art in terms of particular activities, and includes more than 16,000 images of people. We used the workout class videos as a data source and collected images and image sequences using queries based on the descriptions of more than 200 workout activity types. For each activity type, there are 3 different shot angles. This results in a diverse set of images covering not only different workout activities, but contrasting postures. This allows us to enhance the current human pose estimation methods.

\section{Related Work}
There are several human keypoints datasets presented in the past decades. Buffy dataset \cite{ferrari2008progressive} and PASCAL stickmen dataset \cite{eichner2009better} only contain upper-bodies, but we need to process the full-body. In these two datasets pose variation is insignificant. The contrast of image frames is relatively low in the Buffy dataset.

The UIUC people dataset \cite{tran2010improved} contains 593 images (346 for training, 247 for testing). Most people in the images are playing badminton. Some people are playing jogging, Frisbee, standing, walking, etc. There are very aggressive pose and spatial variations. However, the activity type is limited in this dataset.

The sport categories of Sport image dataset \cite{wang2011learning} is more plentiful, which including soccer, cycling, acrobatics, American football, croquet, golf, horseback riding, hockey, figure skating, etc. The total number of images is 1299 (649 of them are split as training set and the rest as testing set).

Leeds Sports Poses (LSP) dataset \cite{johnson2010clustered} includes 2000 images, where one half for training and the other half for testing. The dataset shows people involved in various sports.

The image parsing (IP) dataset \cite{ramanan2006learning} is a small dataset and contains 305 images of fully visible people, where 100 images for training and 205 images for testing. The dataset consists of various activities such as dancing, sports and acrobatics.

The MPII Human pose dataset \cite{andriluka20142d} consists of 24,589 images, in which 17,408 images with 28,883 annotated people are split for training. During the testing stage, one image may contain multiple different evaluation regions that consist of a non-identical number of people. \cite{pishchulin2016deepcut} defines a set of 1,758 evaluation regions on the test images with rough position and scale information. The evaluation metric deploys mean Average Precision (mAP) of the whole body joint prediction. The accuracy results are evaluated and returned by the staff members of the MPII dataset.

The MS-COCO keypoints dataset \cite{lin2014microsoft} includes training, validation and testing sets. On the COCO 2017 keypoints challenge, training and validation sets have 118,287 and 5000 images respectively, totally containing over 150,000 people with around 1.7 million labelled keypoints. In experiments, we perform ablation studies on the validation set. To analyse the effect of training, we also combine the COCO train set with the FollowMeUp train set to validate that new images will not affect the model's generality performance.

The DensePose-COCO dataset \cite{alp2018densepose} has reannotated dense body surface annotations on the 50k COCO images. These dense body surface annotations can be understood as continuous part labels of each human body.

The PoseTrack dataset \cite{andriluka2018posetrack} includes both multi-person pose estimation and tracking annotations in videos. It can perform not only pose estimation in single frames, but also temporal tracking across frames. The dataset contains 514 videos including 66,374 frames in total. The annotation format defined 15 body keypoints. For the single-frame pose estimation, the evaluation metric uses mean average precision (mAP) as is done in \cite{pishchulin2016deepcut}.

\section{The Dataset}

\subsection{Pose Estimation}
The key motivation directing our data selection strategy is that we want to represent rare human postures that might be not easily accessed or captured. To this end, we follow the method of \cite{ainsworth20112011} to propose a two-level hierarchy of workout activities to guide the collection process. This hierarchy was designed according to the body part to be trained during the exercise. The first level is the body part interested to be trained, such as shoulder, whereas the second level is specific workout activities that can strengthen the muscles of shoulder.

\subsubsection{Data collection} We select candidate workout videos according to the hierarchy and filter out videos of low quality and those that people are truncated. This resulted in over 600 videos spanning over 200 different workout types with three shot angles. We also filter out the frames in which pose is not recognisable due to poor image quality, small scale and dense crowds. This step resulted to a total of 110,000 extracted frames from all collected videos. Secondly, since different exercises have disparate periods, we manually pick key frames with people from each video. We aim to select frames that either depict the whole one exercise period in a substantially different pose or different people with dissimilar appearance. The repeated or no significant distinction postures are ignored. Following this step we annotate 16,519 images. We rough randomly split the annotated images for training and use the rest for testing. Images from the same video are either all in the training or all in the test set. We last obtain the train set of 15,435 images and test set of 1,084 images.

\subsubsection{Data annotation} We follow the keypoint annotation format of COCO dataset, where 17 body keypoints are defined. This design facilitates us to utilise the common samples of COCO dataset during training. Following \cite{lin2014microsoft} the left/right joints in the annotations refer to the left/right limbs of the person. Additionally, for all body joints the corresponding visibility is annotated. At test time both the accuracy of joints localisation of a person along with the correct match to the left/right limbs are evaluated. The annotations are performed by in-house workers and inspected by authors. For some unqualified and incorrect annotations are modified continuously until totally correct. To maintain the quality of annotations, we arranged a number of annotation training classes for all annotation workers to unify the standard of annotation. We also supervise and handle some uncertain cases for workers during annotation.


\subsubsection{Pose Estimation Evaluation Metrics} Some previous keypoints evaluation metrics rely on the calculation of body limbs' length, such as ${\rm PCP}$, ${\rm PCK}$ and ${\rm PCK_h}$ used in \cite{andriluka20142d}. However, the workout activities usually have specific postures where the limb's length may be near 0 if the limb is perpendicular to the image plane and the evaluation is not numeric stable in these cases. Therefore comparing the distance between points of groundtruth and prediction directly is more sensible. Here we follow the COCO keypoints dataset, using 5 metrics to describe the performance of a model. They are AP (i.e. average precision), ${\rm AP^0{}^.{}^5}$, ${\rm AP^0{}^.{}^7{}^5}$, ${\rm AP^M}$, ${\rm AP^L}$, as illustrated in Table \ref{tab:cocoMetric}. In the matching between predictions to groundtruth, a matching criterion called object keypoint similarity (OKS) is defined to compute the overlapping ratio between groundtruth and predictions in terms of point distribution \cite{lin2014microsoft}. If OKS is larger than one threshold value (e.g. 0.5), the corresponding groundtruth and prediction are considered as a matching pair and the correctness of predicted keypoint types is further analysed. Here OKS is similar to the intersection over union (IoU) in the case of object detection. Thresholding the OKS adjusts the matching criterion. Notice that in general applications, ${\rm AP^0{}^.{}^5}$ gives a good accuracy already. When computing AP (averaged across all 10 OKS thresholds), 6 thresholds exceed 0.70 are over strict due to unavoidable jittering in annotations.

\begin{table}[!t]
\caption{\label{tab:cocoMetric}Evaluation metrics on the COCO dataset.}
\centering
\begin{tabular}{|p{1.2cm}|p{7.2cm}|}
\hline
Metric & Description\\
\hline
AP & AP at OKS$^*$ = 0.50 : 0.05 : 0.95 (primary metric)\\
\hline
${\rm AP^0{}^.{}^5}$ & AP at OKS = 0.50\\
\hline
${\rm AP^0{}^.{}^7{}^5}$ & AP at OKS = 0.75\\
\hline
${\rm AP^M}$ & AP for medium objects: $32^2 < area < 96^2$\\
\hline
${\rm AP^L}$ & AP for large objects: $area > 96^2$\\
\hline
\multicolumn{2}{@{}l}{$^*$OKS--Object Keypoint Similarity, same role as IoU}
\end{tabular}
\end{table}

\section{Analysis of The State of The Art}
In this section we first compare the leading human pose estimation methods on the COCO keypoints dataset, and then analyse the performance of these approaches on our benchmark.

The basis of the comparison is that we note that there is no uniform evaluation protocol to measure the performance of existing methods from a view of practical application. Although human pose estimation is one of the longest-lasting topics, and significant performance improvement has been achieved in the past few years, some reported accuracies in these approaches are obtained through several post-processing steps or some strategies used in the dataset challenge. For example, performing multi-scale evaluation, refining results by a different method, or precision is evaluated at one image scale while speed is recorded at another scale. These post-processing steps interfere the judgement in identifying the strength and weakness of an algorithm. Therefore, evaluating a method without any post-processing steps and strategies is more objective and more valuable for the research and practical application.

The aim of the analysis is to evaluate the generality of the current models on the different datasets and their performance to the unseen samples, identify the existing limitations and stimulate further research advances.

Currently, there are two main categories of solutions: top-down methods \cite{sun2019deep}\cite{fang2017rmpe}\cite{papandreou2017towards}\cite{xiao2018simple}\cite{chen2018cascaded}\cite{su2019multi} and bottom-up methods \cite{cao2017realtime}\cite{osokin2018real}\cite{insafutdinov2016deepercut}\cite{newell2017associative}\cite{papandreou2018personlab}\cite{kreiss2019pifpaf}. Top-down methods can be seen as a two-stage pipeline from global (i.e. the bounding box) to local (i.e. joints). The first stage is to perform human detection and to obtain their respective bounding boxes in the image. The second stage is to perform single person pose estimation for each of the obtained human regions. \cite{sun2019deep} deploys multiple high-to-low resolution subnetworks with repeated information exchange across multi-resolution subnetworks. This design obtains rich high-resolution representations and leading more accurate result. \cite{fang2017rmpe} utilises a Symmetric Spatial Transformer Network to handle inaccurate bounding boxes. \cite{xiao2018simple} uses simple deconvolution layers to obtain high-resolution heatmaps for human pose estimation. On the side of bottom-up methods, \cite{cao2017realtime} proposes a limb descriptor and an efficient bottom-up grouping approach to associate neighbouring joints. \cite{osokin2018real} modifies the network architecture of \cite{cao2017realtime} and optimises the post-processing steps to achieve real-time speed on the CPU devices. \cite{kreiss2019pifpaf} designs two new descriptors based on \cite{cao2017realtime} for body joints and limbs with the additional variable of object's spread. \cite{newell2017associative} presents a network to simultaneously output keypoint detections and the corresponding keypoint group assignments. \cite{moon2018posefix} designs a feedback architecture that combining the keypoint results of other pose estimation methods with the original image as the new input to the human pose estimation network. In our analysis we consider 8 state-of-the-art multi-person pose estimation methods, which are listed in Table \ref{tab:compare_coco}.

We compare the performance of each approach in terms of accuracy and speed on the COCO dataset and our novel FollowMeUp dataset. All the experiments are performed on a desktop with one NVIDIA GeForce GTX-2080Ti GPU. Since all testing approaches are trained and optimised on the COCO dataset, their open source codes have the corresponding configurations, we directly use their default parameters in our testing.

\subsection{Comparisons of Approaches on the COCO Dataset}
Table \ref{tab:compare_coco} presents the comparison results of testing approaches on the COCO dataset. The upper part of Table \ref{tab:compare_coco} are top-down approaches. \cite{sun2019deep} has the highest AP precision of 0.753. Note that the runtime costs around 50 ms as this only includes the part of pose estimation since this open source library uses the groundtruth of human bounding box as the human detection results on the COCO validation set. \cite{xiao2018simple} and \cite{fang2017rmpe} have a relatively lower accuracy than \cite{sun2019deep} using smaller input sizes, which illustrates that the high-resolution and detailed representation is important for the task of human pose estimation. Note that some post-processing strategies, such as multi-scale and flip, are ignored to obtain the actual performance in the real application environments.

\begin{figure}[t]
\centering
\includegraphics[scale=0.6]{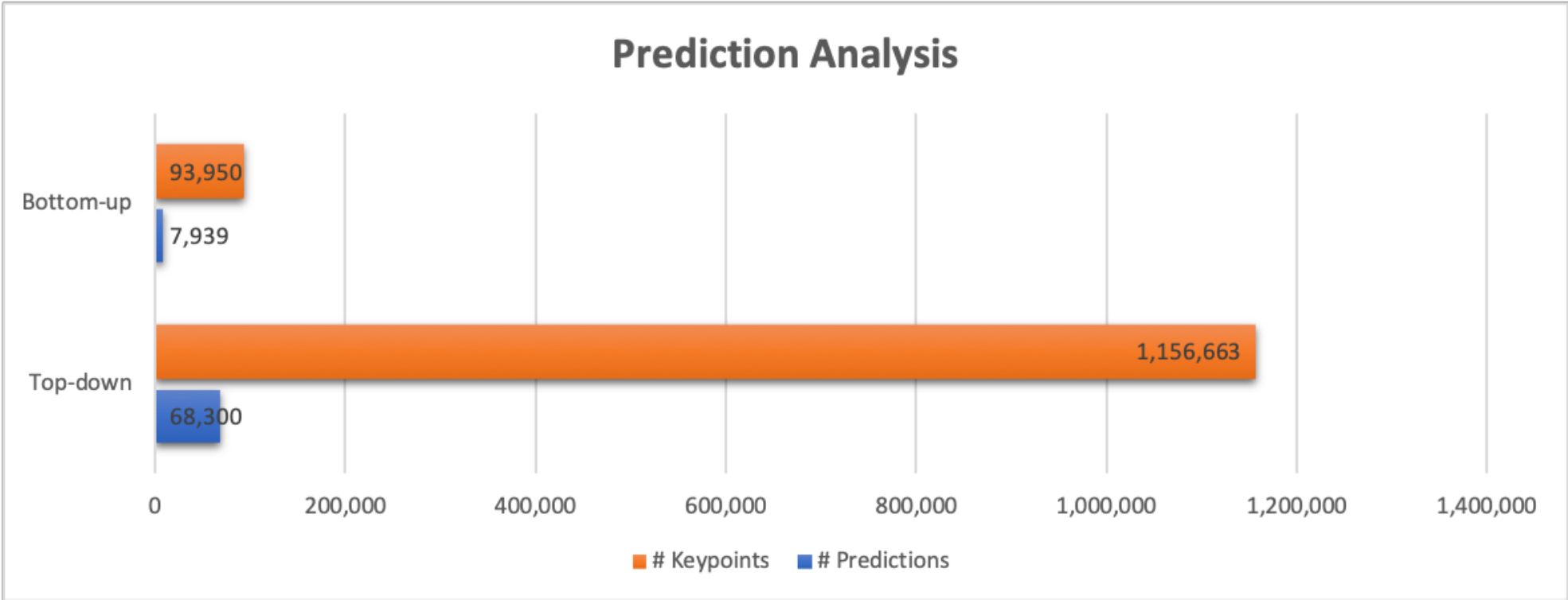}
\caption{The comparison of the numbers of effective instance predictions and body keypoints between top-down and bottom-up methods. The prediction number of top-down method is around 10 times higher than bottom-up method.}
\label{figPredAnalysis}
\end{figure}

For the bottom-up methods, \cite{cao2017realtime} achieves the fastest speed. \cite{kreiss2019pifpaf} attains the highest precision in this group. The joint grouping part of \cite{kreiss2019pifpaf} costs much longer time than \cite{cao2017realtime}. \cite{osokin2018real} has around 7\% degradation compared with \cite{cao2017realtime} due to using a light-weight network architecture. We also see that the precision of bottom-up algorithms are lower than top-down methods. After detailed analysis, we find that the numbers of predicted effective keypoints of bottom-up methods are around 10 times less than top-down methods as illustrated in Fig. \ref{figPredAnalysis}.
We note that top-down methods correspond to performing single-person pose estimation on each detected human region. Single-person pose estimation can output all types of keypoints even the keypoint is occluded or truncated. However, for multi-person bottom-up methods, two or more overlapping keypoints with the same type can only be detected one due to depth information is not available on the RGB image. For the COCO dataset, there are a lot of crowded and occluded human instances. Therefore, the performance of bottom-up methods is weakened. In the FollowMeUp dataset, the crowding case is rare while most human instances have self-occlusion. We perform the same comparison on the FollowMeUp dataset and validate that bottom-up methods have comparable performance to top-down approaches in this circumstance.

\begin{table*}[!t]
\caption{\label{tab:compare_coco}Comparisons of pose estimation results on the COCO 2017 validation set.}
\centering
\begin{tabular}{|p{1.6cm}|p{1.3cm}|p{0.8cm}|p{0.9cm}|p{1cm}|p{0.8cm}|p{0.8cm}|p{1.5cm}|p{1.3cm}|}
\hline
Type & Method & AP & ${\rm AP^0{}^.{}^5}$ & ${\rm AP^0{}^.{}^7{}^5}$ & ${\rm AP^M}$ & ${\rm AP^L}$ & Input Size & Runtime \\
\hline
Top-down & HRNet\cite{sun2019deep} & \textbf{0.753} & \textbf{0.925} & \textbf{0.825} & \textbf{0.723} & \textbf{0.803} & 384x288 & 0.049$^*$ \\
 & Xiao\cite{xiao2018simple} & 0.723 & 0.915 & 0.803 & 0.695 & 0.768 & 256x192 & 0.110 \\
 & RMPE\cite{fang2017rmpe} & 0.735 & 0.887 & 0.802 & 0.693 & 0.799 & 320x256 & 0.298 \\
\hline
Bottom-up & PAF\cite{cao2017realtime} & 0.469 & 0.737 & 0.493 & 0.403 & 0.561 & 432x368 & \textbf{0.081} \\
 & Osokin\cite{osokin2018real} & 0.400 & 0.659 & 0.407 & 0.338 & 0.494 & 368x368 & 0.481\\
 & PifPaf\cite{kreiss2019pifpaf} & 0.630 & 0.855 & 0.691 & 0.603 & 0.677 & 401x401 & 0.202\\
 & AE\cite{newell2017associative} & 0.566 & 0.818 & 0.618 & 0.498 & 0.670 & \textbf{512x512} & 0.260\\
 & PoseFix\cite{moon2018posefix} & 0.411 & 0.647 & 0.412 & 0.303 & 0.559 & 384x288 & 0.250 \\
\hline
\multicolumn{2}{@{}l}{$^*$: without human detection}
\end{tabular}
\end{table*}

\subsection{Comparisons of Approaches on the FollowMeUp Dataset}
Table \ref{tab:compare_followmeup} provides the comparison results of testing approaches on the COCO dataset. Since the open source libraries of \cite{sun2019deep} and \cite{xiao2018simple} do not provide default human detection algorithm, using different human detector may bias the precision distribution, thus we do not test \cite{sun2019deep} and \cite{xiao2018simple} on the FollowMeUp dataset. We are surprised that \cite{fang2017rmpe} obtains a very high precision value. However, the training set only including the COCO dataset of \cite{cao2017realtime} just achieve the precision of 0.778. We argue that the training set of \cite{fang2017rmpe} may include other samples except the COCO dataset with particular postures. In this dataset, the precision of \cite{osokin2018real} decreases by 13\% in ${\rm AP^0{}^.{}^5}$ compared with \cite{cao2017realtime}, which indicates that the generality of \cite{osokin2018real} is also narrowed. We use the results of \cite{cao2017realtime} as the initial poses of \cite{moon2018posefix}. Through pose refinement, \cite{moon2018posefix} improved the pose estimation results by 0.4\%.

\begin{table*}[!t]
\caption{\label{tab:compare_followmeup}Comparisons of pose estimation results on the FollowMeUp dataset.}
\centering
\begin{tabular}{|p{1.6cm}|p{1.7cm}|p{1cm}|p{1cm}|p{1cm}|p{1cm}|p{1cm}|}
\hline
Type & Method & ${\rm AP^0{}^.{}^5}$ & ${\rm AP^0{}^.{}^6}$ & ${\rm AP^0{}^.{}^7}$ & ${\rm AP^0{}^.{}^8}$ & ${\rm AP^0{}^.{}^9}$ \\
\hline
Top-down & RMPE\cite{fang2017rmpe} & 0.975 & 0.948 & 0.885 & 0.787 & 0.421 \\
\hline
Bottom-up & PAF\cite{cao2017realtime} & 0.778 & 0.728 & 0.625 & 0.474 & 0.326 \\
 & Osokin\cite{osokin2018real} & 0.645 & 0.585 & 0.520 & 0.370 & 0.215 \\
 & PoseFix\cite{moon2018posefix} & 0.782 & 0.716 & 0.621 & 0.466 & 0.334 \\
\hline
\end{tabular}
\end{table*}

\subsection{The Effect of Training on the FollowMeUp Dataset}
\begin{figure}[t]
\centering
\includegraphics[width=9.5cm]{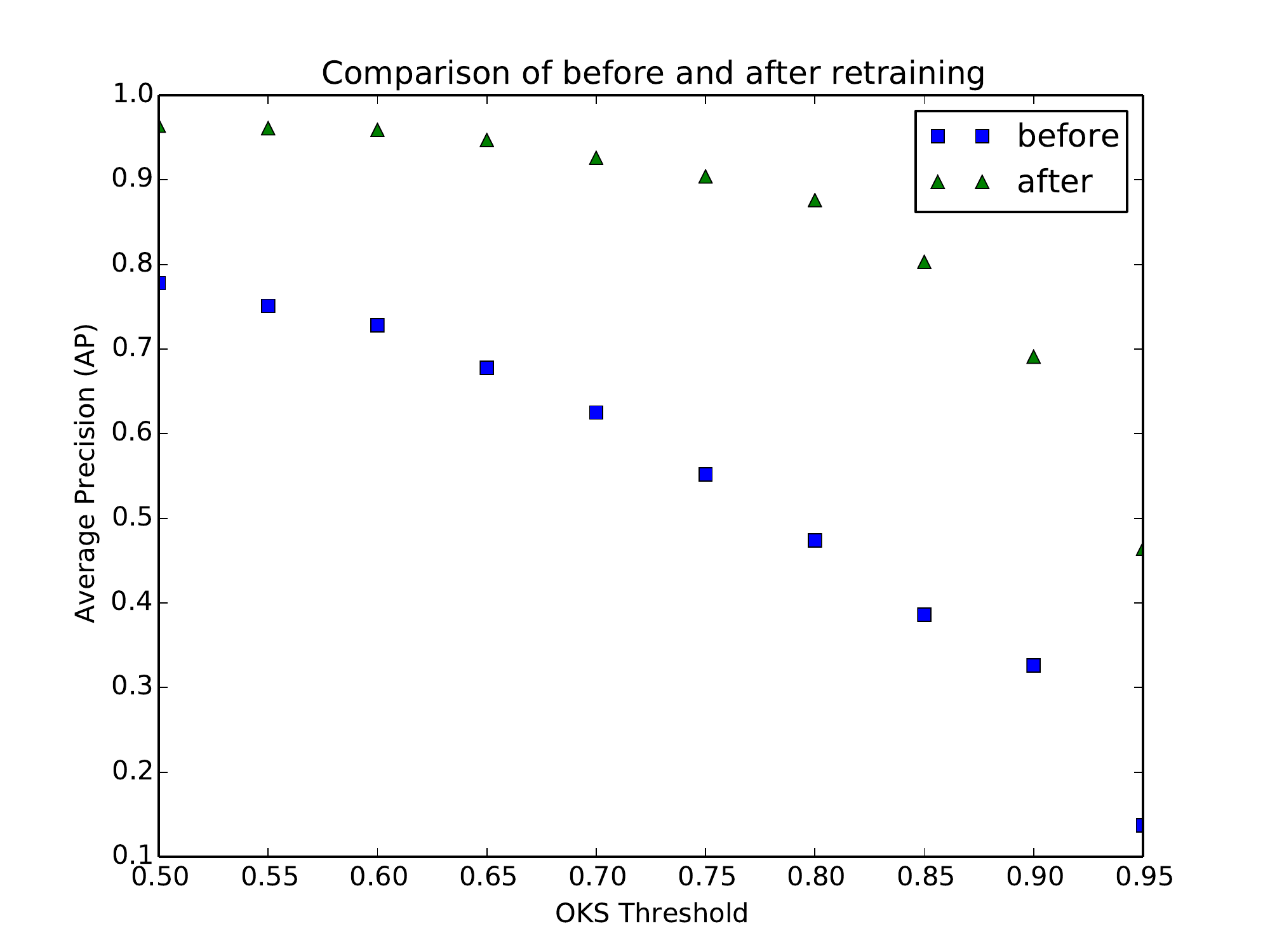}
\caption{Comparison of estimation accuracy before and after retraining on the FollowMeUp dataset. The accuracy of retrained model (marked as green triangles) has an obvious improvement.}
\label{ap_compare}
\end{figure}

To validate the effectiveness of samples with particular postures, we retrain the model on the COCO + FollowMeUp train set using the method of \cite{cao2017realtime}. Testing is performed both on the FollowMeUp test set and COCO validation set. The results of testing are provided in Table \ref{tab:train_followmeup}. We notice that the performance of the retrained model is greatly improved by around 20\% in ${\rm AP^0{}^.{}^5}$. While the threshold of AP becomes more strict, the AP value is decreased. Even in the most strict threshold of 0.9, the AP value attains 0.691, which is higher than the model before retraining by 37\%. The accuracy comparison of before and after retraining on the FollowMeUp dataset is shown in Fig.~\ref{ap_compare}. We also perform testing on the COCO validation set using before and after retraining models to check whether the model can maintain the performance on the COCO dataset. In Table \ref{tab:test_followmeup} we see that before and after retraining the precision has no change. The generality of the retrained model is preserved. These results show that increasing some unusual samples which had not been learnt by the model before is an effective way to improve the accuracy in some specific scenes.

\begin{table*}[!t]
\caption{\label{tab:train_followmeup}Comparisons of pose estimation results on the FollowMeUp dataset.}
\centering
\begin{tabular}{|p{1.3cm}|p{3.2cm}|p{1.8cm}|p{0.8cm}|p{0.8cm}|p{0.9cm}|p{0.8cm}|p{0.8cm}|}
\hline
Method & Train Set & Test Set & ${\rm AP^0{}^.{}^5}$ & ${\rm AP^0{}^.{}^6}$ & ${\rm AP^0{}^.{}^7}$ & ${\rm AP^0{}^.{}^8}$ & ${\rm AP^0{}^.{}^9}$ \\
\hline
PAF\cite{cao2017realtime} & COCO & FollowMeUp & 0.778 & 0.728 & 0.625 & 0.474 & 0.326 \\
PAF\cite{cao2017realtime} & COCO + FollowMeUp & FollowMeUp & 0.964 & 0.959 & 0.926 & 0.876 & 0.691 \\
\hline
\end{tabular}
\end{table*}

\begin{table*}[!t]
\caption{\label{tab:test_followmeup}Comparisons of pose estimation results on the COCO dataset.}
\centering
\begin{tabular}{|p{1.3cm}|p{3.2cm}|p{1.2cm}|p{0.8cm}|p{0.8cm}|p{1cm}|p{0.8cm}|p{0.8cm}|}
\hline
Method & Train Set & Test Set & AP & ${\rm AP^0{}^.{}^5}$ & ${\rm AP^0{}^.{}^7{}^5}$ & ${\rm AP^M}$ & ${\rm AP^L}$ \\
\hline
PAF\cite{cao2017realtime} & COCO & COCO & 0.465 & 0.740 & 0.447 & 0.379 & 0.597 \\
PAF\cite{cao2017realtime} & COCO + FollowMeUp & COCO & 0.465 & 0.748 & 0.454 & 0.373 & 0.605 \\
\hline
\end{tabular}
\end{table*}


\section{Conclusion}
The problem of human pose estimation has obtained a great progress in recent years. This progress cannot be done without the development of large-scale human pose datasets. However, the existing human pose datasets are not sufficient for some particular application environments. In this paper, we propose a new large-scale workout activity human pose dataset, which provides a wide variety of sport exercise postures. We select 8 state-of-the-art multi-person pose estimation approaches and compare their performance on both the popular COCO keypoints dataset and our FollowMeUp dataset. The comparison results show that most methods trained on the COCO dataset do not have ideal performance on the FollowMeUp dataset. We also test the generality of the model using the data of both COCO and FollowMeUp datasets. The test results show that training on the data of both COCO and FollowMeUp datasets will not affect the performance of the model on the COCO dataset but the performance of the model on the FollowMeUp dataset is greatly improved. In the future, we will continue investigate pose tracking\cite{raaj2019efficient}, multi-view action recognition\cite{zhao2019semantic}, and light-weight network design\cite{zf2019fastHumanPose} approaches on the FollowMeUp dataset.

%
%
%
%

\end{document}